\documentclass[10pt,twocolumn,letterpaper]{article}

\usepackage{iccv}
\usepackage{times}
\usepackage{epsfig}
\usepackage{graphicx}
\usepackage{amsmath}
\usepackage{amssymb}
\usepackage{array}

\usepackage{amsmath}

\DeclareMathOperator*{\argmin}{arg\,min}

\usepackage{soul,color}
\usepackage{pifont}
\newcommand{\cmark}{\ding{51}}%
\newcommand{\xmark}{\ding{55}}%
\usepackage[pagebackref=true,breaklinks=true,letterpaper=true,colorlinks,bookmarks=false]{hyperref}

\iccvfinalcopy 

\ificcvfinal \fi

\begin{document}

\title{Meta Learning with Differentiable Closed-form Solver for Fast Video Object Segmentation}

\author{Yu~Liu$^{1}$ 
\and Lingqiao~Liu$^{1} $
\and Haokui~Zhang$^{2}$
\and Hamid~Rezatofighi$^{1}$
\and Ian~Reid$^{1}$
\\
$^{1}$The University of Adelaide, Australia\\
$^{2}$ Northwestern Polytechnical University, China\\
}


\maketitle

\begin{abstract}
This paper tackles the problem of video object segmentation.  We are specifically concerned with the task of segmenting all pixels of a target object in all frames, given the annotation mask in the first frame.  Even when such annotation is available this remains a challenging problem because of the changing appearance and shape of the object over time.
In this paper, we tackle this task by formulating it as a meta-learning problem, where the base learner grasping the semantic scene understanding for a general type of objects, and the meta learner quickly adapting the appearance of the target object with a few examples. Our proposed meta-learning method uses a closed form optimizer, the so-called ``ridge regression", which has been shown to be conducive for fast and better training convergence.
Moreover, we propose a mechanism, named ``block splitting", to further speed up the training process as well as to reduce the number of learning parameters. 
In comparison with the-state-of-the art methods, our proposed framework achieves significant boost up in processing speed, while having very competitive performance compared to the best performing methods on the widely used datasets.


 
\end{abstract}

\begin{figure}[t]
\begin{center}
 \includegraphics[width=1.0\linewidth]{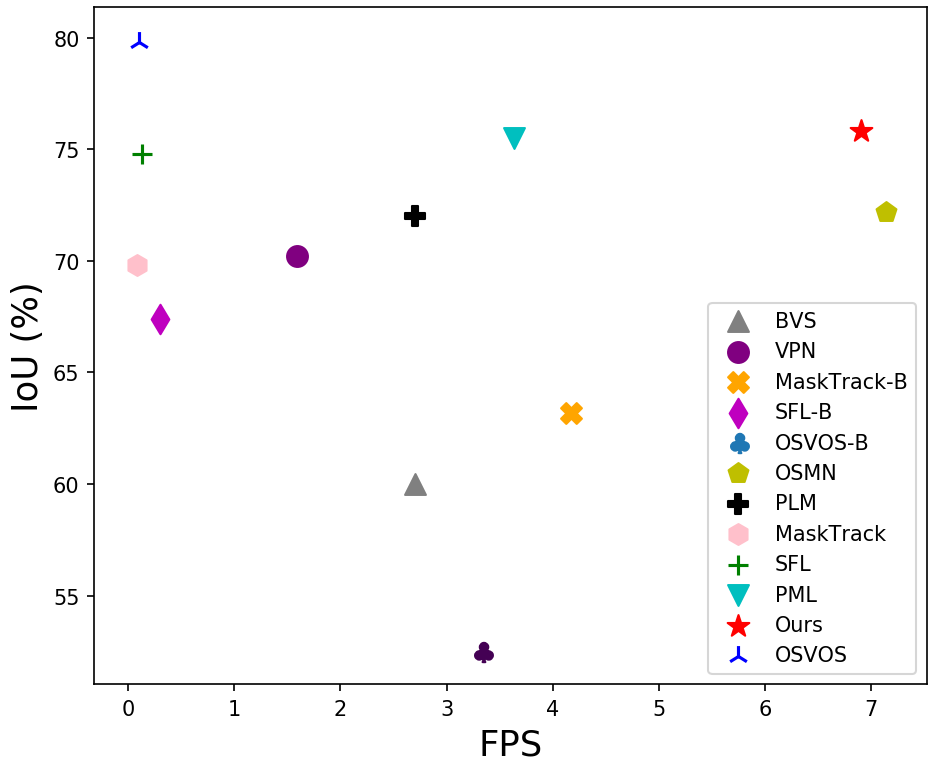}
\end{center}
   \caption{A comparison of the quality and the speed
of previous video object segmentation methods (DAVIS2016
benchmark). We visualize the intersection-over-union
(IoU) with respect to the frames-per-second (FPS).}
\end{figure}
\label{fig: meta_teaser}

\section{Introduction}
Fast and accurate video object segmentation plays an important role in many real-world applications, including, but not limited to, film making~\cite{gee2009deep}, public surveillance~\cite{xu2016video}, robotic vision~\cite{lenz2015deep}. 
\begin{figure*}[t]
\begin{center}
 \includegraphics[width=1.0\linewidth]{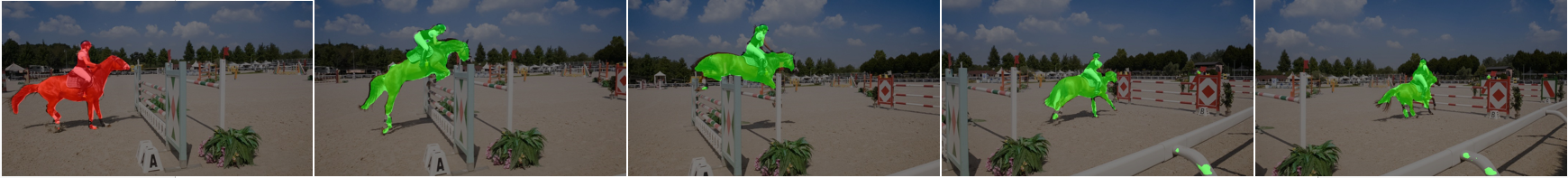}
 
\end{center}
   \caption{Example result of our technique: The segmentation of the first frame (red) is used to learn the model of the specific object to
track, which is segmented in the rest of the frames independently (green). One every 10 frames shown of 50 in total.}
\end{figure*}
\label{fig: teaser}
The goal of video object segmentation is to distinguish an object of interest over video frames from its background at the pixel level. 

In contrast to many vision tasks such as image classification~\cite{NIPS2012_4824}, face recognition~\cite{parkhi2015deep} and object detection~\cite{Redmon_2016_CVPR,Girshick_2015_ICCV} which the performance of the algorithms reach to the point of being suitable for real-world applications, the performance of video object segmentation algorithms are still far beyond the annotations performed by human~\cite{Perazzi_2016_CVPR}. This is mainly because this problem does not benefit from availability of a massive corpus of training data, unlike the other aforementioned tasks.

Recently, deep learning-based approaches have shown promising progresses on video object segmentation task~\cite{caelles2017one,voigtlaender2017online,maninis2017video,sun2018mask}. However, they still struggle to satisfy both good accuracy and fast processing inference. In this paper, we aim to bridge this gap.

\begin{figure*}[t]
\begin{center}
 \includegraphics[width=1.0\linewidth]{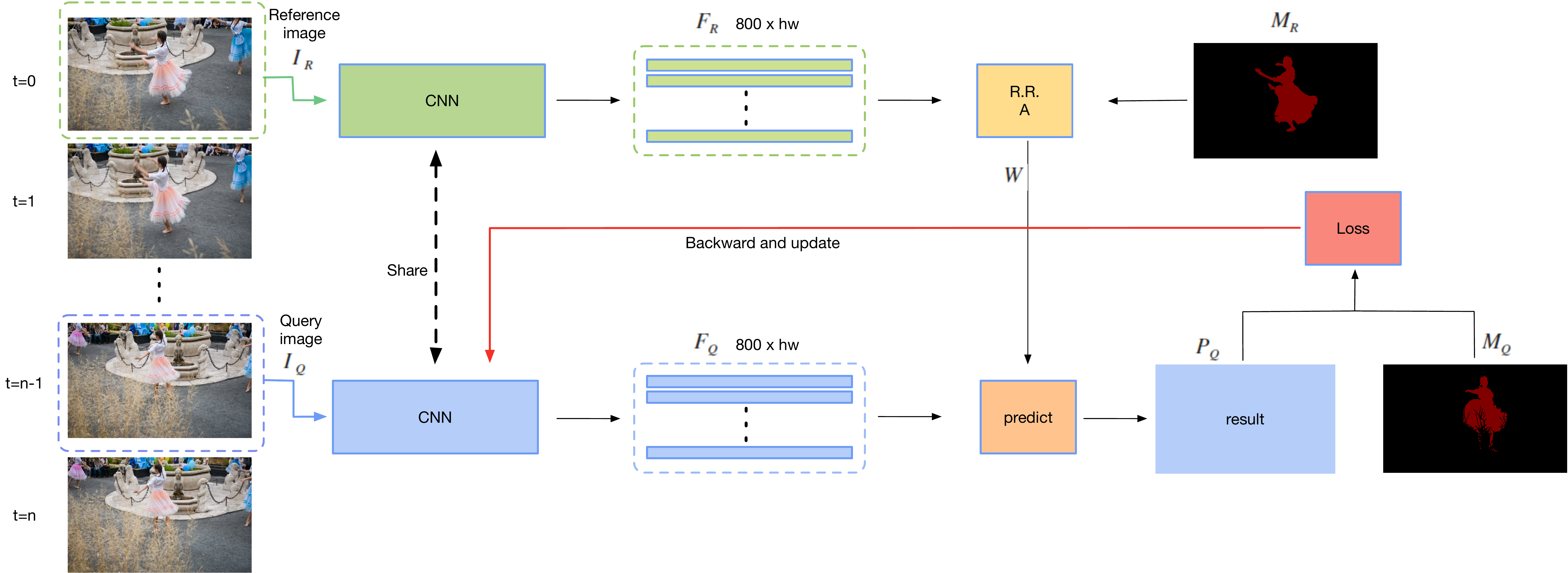}
 
\end{center}
   \caption{Workflow of the proposed method. An image pair sampled from the same video as the input to the network. The first image $I_{R}$ and its annotation
   $M_{R}$ as the reference frame, and the second image $I_{Q}$ and its annotation $M_{Q}$ ( or prediction $P_{Q}$ during inference) as the query frame. The image pair first passes through the \textit{feature extractor} (DeepLabv2~\cite{chen2018deeplab} with ResNet101~\cite{he2016deep}) to compute a 800D embedding tensor $F_R, F_Q$. Then a \textit{mapping matrix W} between
   $F_R$ and $M_R$
   is calculated in the reference frame (Eq. 1) using ridge regression. After that, the prediction result $P_Q$ in the query frame is acquired by multiplying $F_Q$ and $W$ (Eq. 2). During training, the loss error between $P_Q$ and $M_Q$ is back-propagated to enhance the network' adaptation ability between the reference frame and the query frame. During inference, the reference frame ($I_R$ and $M_R$) is always the first frame, and the query image $I_Q$ is the rest sequence from the same video. Through iterative meta-learned, our network is capable of quickly adapting to unseen target object(s) with a few examples. }
\label{fig:workflow}
\end{figure*}

Inspired by the meta-learning method of \cite{bertinetto2018meta}, we propose an intuitive yet powerful algorithm for video object segmentation, in which the reference frame is available with its annotated mask. Our objective is to train a system that can ``adapt'' this annotation information to subsequent frames in a fast yet flexible way at inference time. 
Specifically, at inference time the reference frame (i.e. one with ground-truth annotation) is mapped to vector in a high dimensional embedding space $X = \phi(I)$ using a CNN $\phi$. We then determine using \emph{ridge regression}~\cite{myers1990classical}, the coefficients of a matrix $W$ that best maps $X$ to the ground truth, $Y=WX$. $W$ is then the video-specific ``adaptor'', and it maps the feature vectors for every query image (\ie every other image in the video sequence) to their predicted segmentation masks.  Training comprises the process of learning the mapping $\phi(.)$ by presenting the network with pairs of images (from a variety of videos but with each pair coming from the same video), each with ground-truth annotation, and back-propagating the loss through $\phi$. This is illustrated in Figure \ref{fig:workflow} and described in more detail later in the paper. 

We observe that a limitation of the proposed approach is that the ridge regression scales poorly with the dimension of the feature feature produced by $\phi(.)$ because the optimization requires an huge matrix inversion. We address this through the use of a ``block splitting'' method that approximates the matrix in block diagonal form, meaning the inversion can be done much more efficiently. 


Our main contributions are three-fold:

\begin{itemize}
\item  A meta-learning based method for video object segmentation is developed, using a closed form solver (ridge regression) as the internal optimizer. This is capable of performing fast gradient back-propagation and can adapt to previously unseen objects quickly with very few samples. Inference (i.e. segmentation of the video) is a single forward pass per frame with no need for fine-tuning or post-processing. 

\item
Ridge regression in high-dimensional feature spaces can be very slow, because of the need to invert a large matrix. We address this using a novel block splitting mechanism which we show greatly speeds the training process without damaging the performance.


\item We demonstrate state-of-the-art video segmentation accuracy relative to all others methods of comparable processing time, and even better accuracy than many slower ones (see Figure \ref{fig: teaser}).
\end{itemize}


\section{Related Works}
\label{relatedworks}
\subsection{Semi-supervised Video Object Segmentation}

The goal of video object segmentation is to `cutout' the target object(s) from the entire input video sequence. Regarding the amount of supervision utilized for video object segmentation, methods can be roughly put into two spectrum, i.e. semi-supervised and unsupervised methods.

For semi-supervised video object segmentation, the annotated mask of the first frame is given, and the algorithm is designed to predict the masks of the rest frames in the video. There are three categories in this spectrum. The first one, which include  MSK~\cite{perazzi2017learning}, MPNVOS~\cite{sun2018mask} etc, is to use optical flow to track the mask from the previous frame to the current frame. Similarly, the second category formulates the optical flow and segmentation in two parallel branches, and utilizes the predicted mask from the previous frame as a guidance, some representatives are 
Segflow~\cite{cheng2017segflow},VSOF~\cite{Tsai_2016_CVPR}, RGMP~\cite{wug2018fast}, OSNM~\cite{yang2018efficient} etc. The final class which keeps the state-of-the-art performance on Davis benchmark~\cite{Perazzi_2016_CVPR} is to try to over-fit the appearance of the target object(s), and expect the method can generalize in the subsequent frames. Specifically, OSVOS~\cite{caelles2017one} uses one-shot learning mechanism to conduct fine-tuning on the first frame of test video to capture the appearance of the target object(s), and conduct inference on the rest frames. The drawbacks of OSVOS are: (1) it can not adapt to the unseen parts (2) when dramatic changes of appearance happen in subsequent frames, the method's performance significantly degrade. Inspired by the overall design of OSVOS, there are some following methods which employ various additional ingredients to improve the segmentation accuracy.
In particular, OSVOS-S~\cite{maninis2017video} combines the semantic instance information to remove the noisy objects coming from the same category. OnVOS~\cite{voigtlaender2017online} utilizes on-line adaption mechanism to overcome the limits of OSVOS when drastic appearance changes occur. CINM~\cite{bao2018cnn} utilizes a CNN-based markov random field (MRF) to estimate the probabilities of the pixels belonging to the target object(s) in spatial domain, and employs optical flow to track segmented pixels in temporal domain. Albeit those methods improve the segmentation performance of OSVOS to some extent, they are still time-consuming during inference since the on-line fine-tuning is necessary. And usually, at lease a dense CRF~\cite{krahenbuhl2011efficient} and more techniques are applied as the post-processing step to acquire the better segmentation results.


In this paper, we mainly target to fast video object segmentation, since no optical flow and fine-tuning processes are used, the proposed method is appropriate for real-world applications.

\subsection{Meta Learning}
Meta learning is also named learning to learn~\cite{schmidhuber1987evolutionary,naik1992meta,thrun2012learning} because its goal is to help the machine to be capable of learning quickly, especially in the case with very few samples for the new task(s). Generally speaking, meta learning algorithms are composed of two components, i.e. \textit{base learner} and \textit{meta learner}. According to their roles, base learner is mainly in charge of handling with individual tasks, and meta learner is much like a coordinator, through learning individuals tasks, meta learner can boost the performance of base learners across the tasks.

Meta-learning is an alternative to the de-facto solution that has emerged in deep learning of pre-training a network using a large, generic dataset (eg ImageNet~\cite{deng2009imagenet}) followed by fine-tuning with a problem-specific dataset. Meta-learning aims to replace the fine-tuning stage (which can still be very expensive) by training a network that has a degree of plasticity so that it can adapt rapidly to new tasks. For this reason it has become a very active area recently, especially with regard to one-shot and few-shot learning problems~\cite{lake2015human,fei2006one} 

Recent approaches for meta-learning can be roughly put into three categories: (i) metric learning for acquiring similarities~\cite{vinyals2016matching,snell2017prototypical,garcia2017few}; (ii) learning optimizers for gaining update rules~\cite{finn2017model,ravi2016optimization}; and (iii) recurrent networks for reserving the memory~\cite{santoro2016meta,kaiser2017learning}. In this work, we adopt the meta-learning algorithm that belongs to the category of learning optimizers. Specifically, inspired by~\cite{bertinetto2018meta} which was originally designed for image classification, we adopt ridge regression, which is a closed-form solution to the optimization problem. The reason for using it is because, compared with the widely-used SGD~\cite{lecun1998gradient} in CNNs, ridge regression can propagate gradient efficiently, which is matched with the goal of \textit{fast mapping}.
Through extensive experiments, we demonstrate that the proposed method is in the first echelon
regarding to speed for fast video object segmentation, while obtaining more accurate results without any post-processing.

\subsection{Fast Video Object Segmentation}
A few previous methods proposed to tackle fast video object segmentation. In particular, FAVOS~\cite{cheng2018fast} first tracks the part-based detection. Then, based on the tracked box, it generates the part-based segments and merges those parts according to a similarity score to form the final segmentation results. The drawback of FAVOS is that it can not be learned in an end-to-end manner, and heavily relies on the part-based detection performance. OSNM~\cite{yang2018efficient} proposes a model which is composed of 
a modulator and a segmentation network. Through encoding the mask prior, the modular can help the segmentation network quickly adapt to the target object. RGMP~\cite{wug2018fast} 
shares the same spirit with OSNM. Specifically, it employs a Siamese encoder-decoder structure to utilize the mask propagation, and further boosts the performance with
synthetic data. The most similar work to ours is
PML~\cite{chen2018blazingly}, which formulates the problem as a pixel-wise metric learning problem. Through the FCN~\cite{long2015fully}, it maps the pixels to high-dimensional space, and utilizes a revised triplet loss to encourage pixels belonging to the same object much closer than those belonging to different objects. Nearest neighbor (NN) is required for retrieval during inference. 
In contrast our meta-learning approach 
acquires a mapping matrix between the high-dimensional feature and annotated mask in reference image using ridge regression, and then can be adapted rapidly to generate the prediction mask. Compared to baseline method PML~\cite{chen2018blazingly}, our method achieves more accurate performance and is twice the speed. And with the same efficiency, the \textit{J mean} of our method is 3.4 percent better than OSNM~\cite{yang2018efficient}
on the DAVIS2016~\cite{Perazzi_2016_CVPR} validation set.

\section{Methodology}
\label{methology}
\noindent
\subsection{Overview}
We formulate the video object segmentation as a meta-learning problem. For each image pair which comes from a same video, ridge regression is used as the optimizer to learn the base learner. Meta learner is naturally built through the training process. Once the meta learner is learned, it possesses the ability of \textit{fast mapping} between the image features and object masks, 
and can be adapted to unseen objects quickly with the help of the reference image.

According to the phase that user input involved in the training loop, the current existing methods can be classified into three categories.

\noindent
\textbf{User input outside the network training loop} This category utilizes the user input to fine-tune the network to over-fit the appearance cues of target object(s) during inference. The representatives are OSVOS~\cite{caelles2017one} and its following works~\cite{maninis2017video,bao2018cnn,voigtlaender2017online}. Since online fine-tuning is required during inference, the drawback of these algorithms is time-consuming, which usually take seconds per image, thus is not practical for the real-world applications. 

\noindent
\textbf{User input within the network training loop}
This category of work injects the user input as the additional input for training the network. Through this way, no online fine-tuning is needed. These algorithms incorporate the user input either by using a parallel network or concatenating the image with the user input~\cite{wug2018fast,yang2018efficient}. One drawback of this kind of methods is that the model needs to be recalculated once the user input changes, thus it is not practical for adaptation especially for long videos.

\noindent
\textbf{User input is detached from the network training loop}
In contrast to the previous methods, our algorithm shares the same spirit with PML~\cite{chen2018blazingly} in design. The network and user input are detached, and the user input can be more flexible. Moreover, once the user input is given (for example, the annotation in the reference image), the network can quickly adapt to the target objects without any extra operations.

\subsection{Segmentation as Meta-Learning}

For simplicity, we assume single-object segmentation case, and the annotation of first frame is given as the user input. Note that our method can also be applied for multi-objects and easily extended to other types of user input, e.g., scribble, clicks etc.

We adopt the following notation:

\noindent
$C$ denotes the number of feature channels (in our case 800). \\
$w, h$ denote the spatial resolution of the extracted features (in our case 1/8th of the orginal image size). \\ 
$F_R, F_Q$ are the feature tensors of size $C \times h \times w$ produced by $\phi$ \\ 
$X$ is a flattened tensor of $F_R$ or $F_Q$, with shape $ h\cdot w \times C$ \\
$Y$ is the flattened 
tensor of annotation mask $M_R$ or $M_Q$, with shape $h\cdot w \times 1$ \\
$W$ denotes the \textit{mapping matrix} of size $C \times 1$ between the feature space and annotation mask. 

As noted above, there are two components to the learner: (i) $\phi(.)$ an embedding model that maps images to a high-dimensional feature space, $C \times h \times w$; and (ii) an adaptor $W$ of size $C \times 1$, found using ridge regression, that maps the embedded features to a (flattened) segmentation mask (of size $h \cdot w \times 1$).

\noindent
\textbf{Embedding Model}
We adopt DeeplabV2~\cite{chen2018deeplab} built on the ResNet-101~\cite{he2016deep} backbone structure as our  feature extractor $\phi$. This choice allows a direct comparison of our method with the baseline, PML~\cite{chen2018blazingly}. 
First, we use the pretrained model on COCO~\cite{lin2014microsoft} dataset as the initialization for semantic segmentation. Then the ASPP~\cite{chen2018deeplab} layer for classification is removed and replaced by our video-specific mapping $W$.

\noindent
\textbf{Ridge Regression}
Ridge regression is a closed form solver and widely-used in machine learning community~\cite{saunders1998ridge,nouretdinov2001ridge}.
The learner seeks $W$ that minimizes $\Lambda$ as follows:
\begin{equation} \label{formula1}
\begin{split}
\Lambda(X,Y)& =
\argmin_W||XW-Y||^2+\lambda||W||^2    \\
 & =(X^TX+\lambda I)^{-1} X^T Y
\end{split}
\end{equation}

\noindent where, $X,Y$ and $W$ are as defined above,
and $\lambda$ is a regularization parameter, and set to 5.0 in all of our experiments. As can be seen in Figure~\ref{fig:workflow}, during training, an image pair as well as their annotations are sampled from the same video sequence. The feature \textit{$F_{R}$} extracted from the reference image \textit{$I_{R}$} (in the figure this is the first image) and its annotation \textit{$M_{R}$} will be used to calculate the mapping matrix $W$.
\begin{equation} \label{formula2}
P_Q = F_Q \times W 
\end{equation}
(where we abuse notation and use the unflattened feature tensors for clarity)

For the query image $I_{Q}$, likewise we compute the feature $F_{Q}$, map these to the predicted segmentation mask $P_{Q}$ using Equation~\ref{formula2}  in which $W$ is the matrix computed from the reference image and its ground truth. 
The loss between the prediction mask $P_{Q}$ and the annotation $M_{Q}$ for the query provides the back-propagation signal to improve $\phi$'s ability to produce adaptable features.

During inference in our case, the reference frame $I_R$ will be always the first frame, for which the annotation mask is provided, and the query frames $I_Q$ will be the rest of frames in the same video.

\subsection{Block Splitting}
\label{sec: block splitting}
Thanks to ridge regression, the computation of the \textit{mapping matrix} and gradient back-propagation are already very fast compared with other algorithms, which also focus on video object segmentation. 
\begin{equation} \label{formula3}
F(X)=(X^TX+\lambda I)^{-1}
\end{equation}

During the experiments, we found the higher dimension of the feature used as the input for meta-learning module, the more accurate segmentation results likely be achieved. However, we also observed that the higher dimension of the feature being utilized, the slower of the training process. Specifically, during the computation of \textit{mapping matrix W}, it involves a matrix inverse calculation. as denoted by Equation~\ref{formula3}, which will become the bottleneck of fast propagation when the very high dimensional feature is used.

\begin{figure}[t]
\begin{center}
  \includegraphics[width=1.0\linewidth]{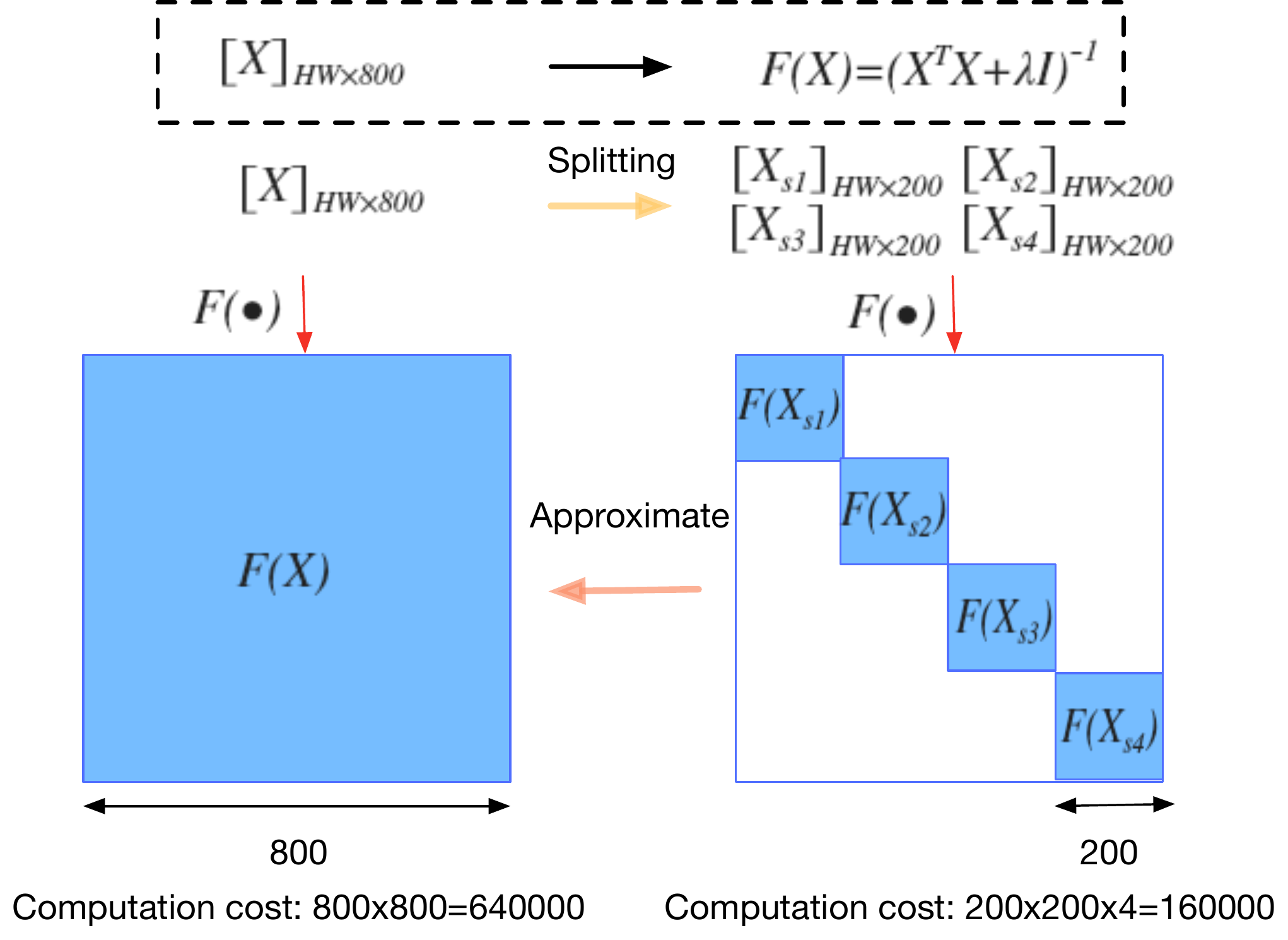}
\end{center}
   \caption{Illustration of the proposed \textit{block splitting}: during matrix inverse calculation of ridge regression, the computation of the higher dimensional feature is approximated by the sum of computation of that lower dimensional features. Which can effectively speed up the training process as well as reducing the parameters and memory. }
\label{fig:vis_block_split}
\end{figure}

In order to further speed up the training process of the proposed network, we deliver a \textit{block splitting} mechanism, and its work principle as shown in Figure~\ref{fig:vis_block_split}. In particular,
, our motivation is that the matrix inverse computation $F(X)$
for much high-dimensional feature (eg. 800D) can be approximated by the sum of the computations of that relative low-dimensional features (eg. 200D $\times$ 4).
From the work principle, it can be viewed that a $n \times n$ matrix can be approximated by four $n/4 \times n/4$ irrelevant diagonal matrix. 

The advantages of using the proposed block splitting mechanism are: Firstly, it can largely speed up the matrix inverse process involved in ridge regression, thus it saves the training time to some extent.
Secondly, through the matrix approximation step as aforementioned, the network parameters involved in the ridge regression as well as memory utilized in our network are reduced. The experimental evidence can be found in Ablation Study ( Section~\ref{ablationstudy}).

\subsection{Training}
\noindent
\textbf{Training Strategy}
For training, optimizer is SGD with momentum 0.9, with weight decay 5e-4. We use the DeepLabV2~\cite{chen2018deeplab} with backbone network ResNet-101~\cite{he2016deep} as the \textit{feature extractor}, and the constant learning rate, i.e. 1.0e-5, is used during the whole training process. The dimension of extracted feature is 800 outputed by the \textit{feature extractor}, which is used as the input for the meta-learning module.

\noindent
\textbf{Loss}
BCEWithLogitsLoss\footnote{https://pytorch.org/docs/stable/nn.html} is employed for training the proposed network, it essentially is a combination of the Sigmoid layer and binary cross entropy (BCE) loss, it benefits from the \textit{log-sum-exp} trick for numerical stability. And compared to BCE loss, it is more robust and less likely to cause numerical problem when computing the inverse matrix in the ridge regression step. 
\label{vis_blocksplit}

\begin{equation}
\begin{matrix}
\ell(x,y)=L=\{l_1,...,l_N\}^T  \\
l_n=-w_n[y_n \cdot \log \delta(x_n)+ (1-y_n) \cdot \log(1-\delta(x_n))]
\end{matrix}
\end{equation}

\noindent
where $N$ is the batch size. $x_n$ is the input of the loss calculation, and $y_n$ ($y_n \in [0,1]$) is the ground truth label. $w_n$ is a rescaling weight given to the loss of each batch element.

\begin{figure*}[t]
\begin{center}
 \includegraphics[width=1.0\linewidth]{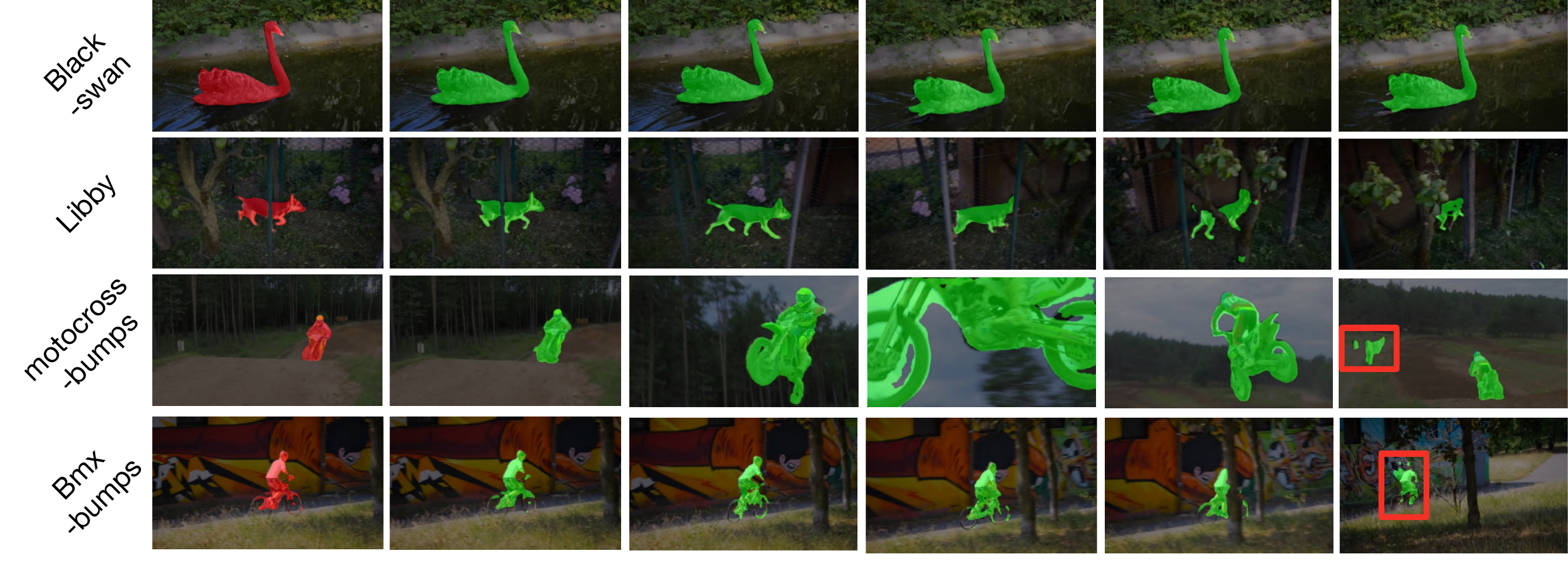} 
\end{center}
   \caption{Qualitative results: Homogeneous sample of DAVIS sequences with our result overlaid}
\label{fig:vis_2016}
\end{figure*}
















\begin{table*}[t]
\centering
\begin{tabular}{m{3cm} | c c c c c c}

\hline
Method & DAVIS16 & Online-Tuning & OptFlow & CRF & BS & Speed(s)\\
\hline
\hline
OFL & 68.0 & - & \xmark & \cmark & \xmark & 42.2 \\

BVS & 60.0 & - & \xmark & \xmark & \xmark & 0.37 \\

ConvGRU & 70.1 & \xmark & \cmark & \xmark & \xmark & 20 \\

VPN & 70.2 & \xmark & \xmark & \xmark & \xmark & 0.63 \\

MaskTrack-B & 63.2 & - & \xmark & \xmark & \xmark & 0.24 \\

SFL-B & 67.4 & \xmark & \cmark & \xmark & \xmark & 0.30 \\

OSVOS-B & 52.5 & \xmark & \xmark & \xmark & \xmark & \textbf{0.14} \\

OSNM & 72.2 & \xmark & \xmark & \xmark & \xmark & \textbf{0.14} \\

PML & 75.5 & \xmark & \xmark & \xmark & \xmark & 0.28 \\

Ours & \textbf{75.8} & \xmark & \xmark & \xmark & \xmark & 0.145 \\
\hline

PLM & 70.0 & \cmark & \xmark & \xmark & \xmark & 0.50 \\

SFL & 74.8 & \cmark & \xmark & \xmark & \xmark & 7.9 \\
MaskTrack & 69.8 & \cmark & \xmark & \xmark & \xmark & 12 \\

OSVOS & \textbf{79.8} & \cmark & \cmark & \xmark & \cmark & 10 \\
\hline
\end{tabular}

\caption{Performance comparison of our approach with recent approaches on DAVIS 2016 Performance measured in
mean IoU.}
\label{tab:addson_2016}
\end{table*}
\section{Experiments}
\label{experiments}

\begin{figure*}[t]
\begin{center}
 \includegraphics[width=0.9\linewidth, height=0.43\linewidth]{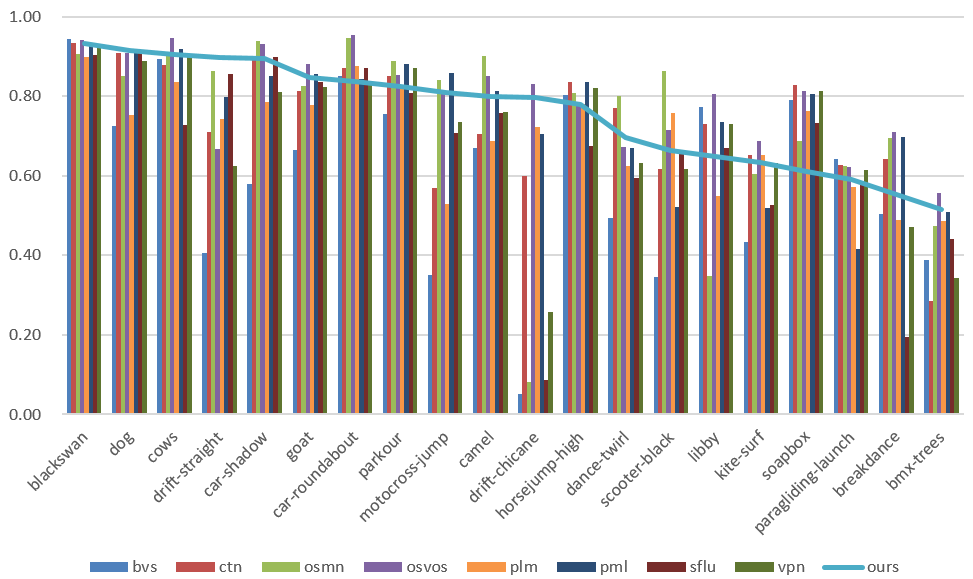}
\end{center}
   \caption{Per-sequence results of mean region similarity (J ) . Sequences are sorted by our performance.}
\label{fig:vis_perseq_davis16}
\end{figure*}
\label{fig:davis2016_perseq}

\subsection{Dataset}
We verify the proposed method both on DAVIS2016~\cite{Perazzi_2016_CVPR} and SegTrack v2~\cite{li2013video} datasets.

On DAVIS2016, which contains 50 pixel-level annotated video sequences, and each video only contains one target object for segmenting. Among these 50 video sequences,
30 video sequences as the training set with which the annotated mask is provided for every frame. And another 20 video sequences as the validation set, and only the annotation of the first frame is allowed to access. 

SegTrack v2~\cite{li2013video} is extended from SegTrack~\cite{tsai2012motion} dataset. Both of them contain the dense pixel-level annotation for each frame within each video. For segtrack v2 dataset, we test our algorithm on all the sequences which contain one target object.
\subsection{Results on DAVIS2016}
\noindent
\textbf{Quantitative Results}
Table~\ref{tab:addson_2016} shows the experimental results on DAVIS2016~\cite{Perazzi_2016_CVPR} on different methods. Apart from the performance (measured by \textit{J mean}), switches for online-fining, using optical-flow, dense CRF (CRF) and boundary snapping (BS) are also described. Meanwhile, the inference time is also shown. In particular, compared with most of the competitors, our algorithm shares the same or much faster processing time with superior performance regarding the segmentation accuracy. Please note that, some methods which use much stronger backbone networks are not listed out for the purpose of fair comparison.
Compared with OSVOS~\cite{caelles2017one}, for which the online fine-tuning is necessary, our method just takes a smaller fraction of time to do inference.
Compared to the baseline method PML ~\cite{chen2018blazingly} which use the same \textit{feature extractor}, our method is twice faster and with better performance. Compared OSNM~\cite{yang2018efficient}, with the same efficiency, our method achieve 3.4 percent improvements regarding to the segmentation accuracy.

\noindent
\textbf{Qualitative Results}
\label{qualitiveRes_davis2016}

Figure~\ref{fig:vis_2016} demonstrates some visualized results of our method. As shown in Figure~\ref{fig:vis_2016}, our method is not only good at recovering object details (e.g., the results on the sequence of \textit{blackswan}), but also robust against heavy occlusions (eg. the results on the sequences  \textit{bmx-bumps} and \textit{libby}, dramatic movement as well as abrupt rotation (eg. the results on the sequence \textit{motocross-bumps}). However, there are very few scenarios which may lead to failure cases (denoted by the red box), and mainly caused by the (noisy) objects which have not appeared at the first frame of the video, and can be easily cured by some post-processing steps, including tracking~\cite{cheng2018fast}, online adaptation~\cite{voigtlaender2017online, chen2018blazingly}.   

\begin{figure}[t]
\begin{center}
 \includegraphics[width=1.0\linewidth]{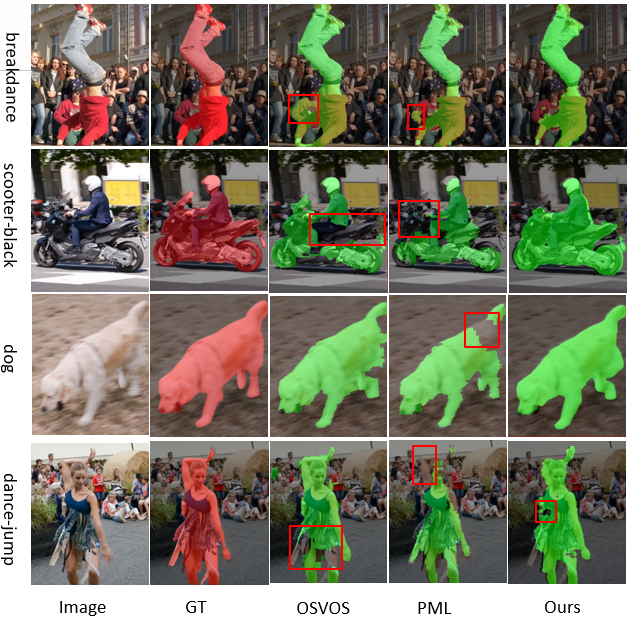} 
\end{center}
   \caption{Visualized comparison between the proposed method and other methods. With the red box to denote the error region.}
\label{fig:vis_compare_2016}
\end{figure}

In Figure~\ref{fig:vis_compare_2016}, we show some visualized results compared with OSVOS~\cite{caelles2017one} and PML~\cite{chen2018blazingly}. For the \textit{breakdance}, \textit{scooter-black} and \textit{dance-jump} sequences, which contain fast moving and abrupt rotation, OSVOS~\cite{caelles2017one} performs worse than PML~\cite{chen2018blazingly}. And for the \textit{dog} sequence, PML~\cite{chen2018blazingly} can not achieve a satisfied result due to the dramatic change of the light conditions. However, on both of these two scenarios, the proposed method performs better than both of OSVOS and PML, which is benefit from robust adaptation ability of our network.

\vspace{-0.29em}
\subsection{Results on SegTrack Dataset}

\begin{figure*}[t]
\begin{center}
 \includegraphics[width=1.0\linewidth]{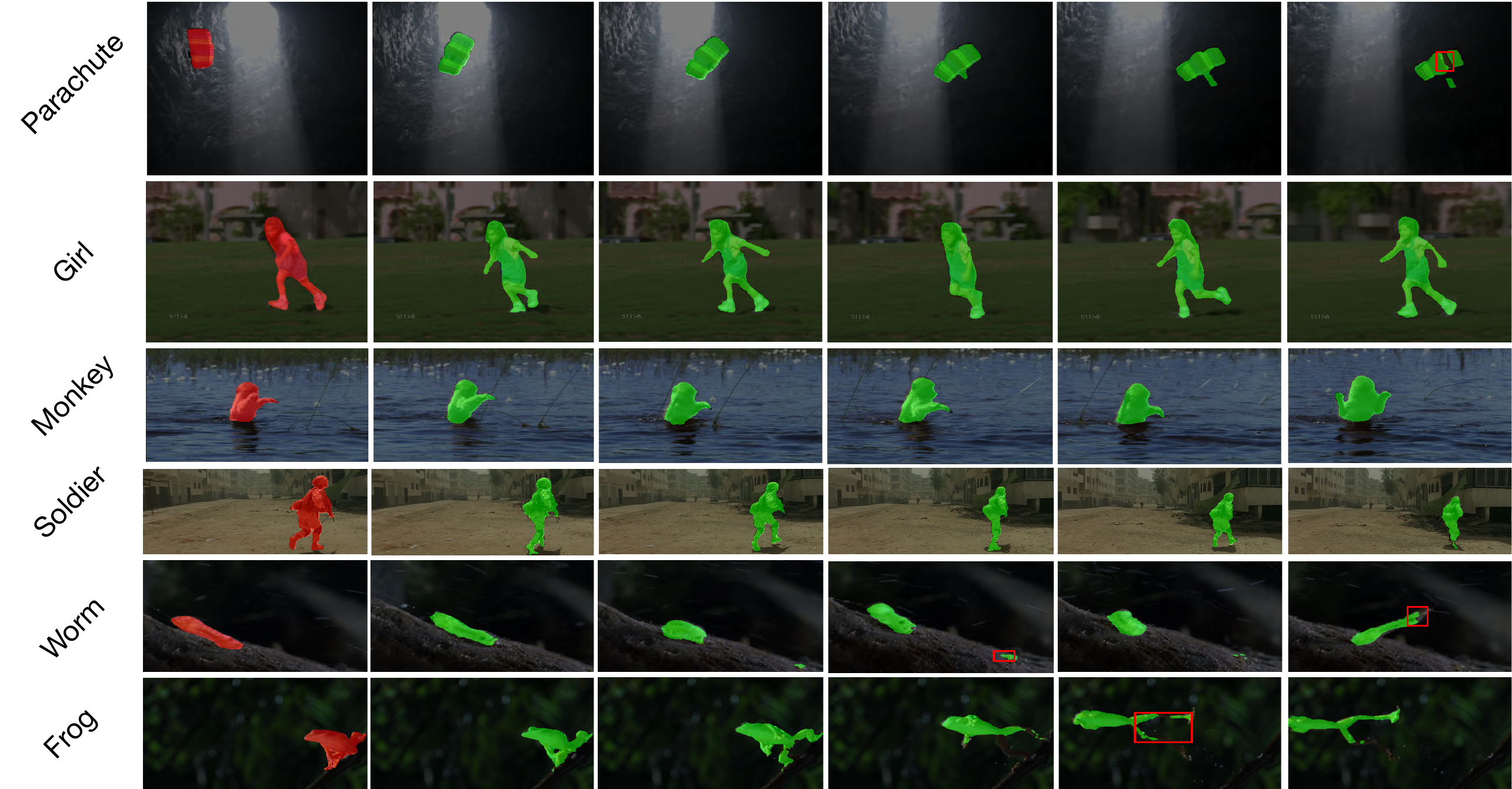}
\end{center}
   \caption{Qualitative results: Homogeneous sample of SegTrack sequences with our result overlaid}
\label{fig:segtrack_vis}
\end{figure*}

In Figure~\ref{fig:segtrack_vis}, some visualized results in the segTrack~\cite{tsai2012motion} dataset are shown. Which are acquired by direcly utlized the model trained on Davis2016 dataset. As can be seen, in most cases, our model maintain a good segmentation accuracy, and with a few case fails (as denoted by the red box), which mainly due to the dramatically changes of the light conditions and exact same appearance between the background and the target object. These results prove our method has a better generalization ability and can be quickly adapted to other unseen objects with very few examples (here, only the annotation in the first frame is provided).

\section{Ablation Study}
\label{ablationstudy}
\subsection{Feature Dimension and Block Splitting}

\begin{table}[t]
\begin{center}
\scalebox{0.9}{
\begin{tabular}{c|c|c|c|c}

Split No& Feature & Speed & Memory & Computation Cost \\

\hline
1 & 800 & 1.50 & 11590&
640k
\\

2 & 400 & 1.23 & 11720 &
320k
\\

4 & 200 & 0.75 & 11580 &
160k
\\

8 & 100 & 0.86 & 11584 &
80k 
\\

\end{tabular}
}
\end{center}
\caption{Ablation study on \textit{block splitting}: feature dimension, running speed, memory and computation cost with different settings are listed out.}
\label{tab:block_split}
\end{table}

As mention in Section~\ref{sec: block splitting}, since our meta learning module (ridge regression) requires the computation of matrix inverse, the training speed will varies significantly regrading the features with various dimensions utilized for this step. And based on the fact that low dimensional features usually have the faster speed but lose some details of image information. On the contrary, high dimensional features are time-consuming but carry much rich information.
We propose a \textit{block splitting} mechanism to train the meta learner. In Table~\ref{tab:block_split}, the splitting number (of feature), feature dimension, running speed (per iteration), memory cost (of the whole network), as well as computation cost (of the computation of matrix inverse) with different settings are listed out. As can be seen, with the feature dimension decreasing, the overall trend are running speed increasing, computation cost decreasing, dramatically. However the memory cost reduce slightly, which mainly because of the backbone \textit{feature extractor} take up most of the memory usage. All the numbers are tested on the single GPU card (with type of GTX 1080). 




\subsection{Per Sequence Performance Analysis}
In Figure~\ref{fig:vis_perseq_davis16}, \textit{J mean} of per sequence of different methods are outlined.
It is sorted according our algorithm's performance in each sub-sequence, which provides a more intuitive understanding for the proposed algorithm. Firstly,
the proposed method achieve a better video segmentation accuracy when compared to many other methods. Secondly, our algorithm works quite well on most of sequences, even on the most challenging sequences, e.g., \textit{breakdance} and \textit{bmx-tree}, the \textit{J mean} is above 0.5.Thirdly, benefit from the quick adaption ability of meta-learning, around half of sequence achieve \textit{J mean} over 0.8. Moreover, our method can well recover the object details as well as robust against fast movement and heavy occlusion, which are aligned with our conclusion in Section~\ref{qualitiveRes_davis2016} 
\section{Conclusion}
\label{conclusion}
In this paper, we explore applying meta-learning into video object segmentation system. A closed form optimizer, i.e., ridge regression, is utilized to update the meta learner, which achieves fast speed while maintains the superior accuracy.
Through iteratively meta-learned, the network is capable of conducting \textit{fast mapping} on unseen objects with a few examples available.
Compared to the fine-tuning methods,
our algorithm with similar performance but just a smaller fraction time is required, which is appeal to the real-world applications. In addition, a block splitting mechanism is delivered to speed up the training process, which also has the benefits of reducing parameters and saving memory.
In future work, we would like to use other basic optimizers, such as, Newton's methods and logistic regression. Meanwhile, based on the flexible design of our meta-learner, instead of inferring the rest frames from the given whole annotation of the first frame. Inferring whole object from only part of annotation or user feedback is also worth to investigate.



\end{document}